\documentclass[9pt,conference]{IEEEtran}
\IEEEoverridecommandlockouts

\usepackage{amsmath,graphicx}
\usepackage{cite}
\usepackage{graphicx}
\usepackage{textcomp}
\usepackage{xcolor}
\usepackage{amssymb}
\usepackage{algorithm,algorithmic}
\usepackage{booktabs}
\usepackage{pifont}       
\usepackage{bbding}       
\usepackage{fontawesome}  
\usepackage{hyperref}

\usepackage{xspace}
\makeatletter
\DeclareRobustCommand\onedot{\futurelet\@let@token\@onedot}
\def\@onedot{\ifx\@let@token.\else.\null\fi\xspace}

\def\etal{\emph{et al}\onedot}
\makeatother
\begin{document}

\title{
Lagrange Duality and Compound Multi-Attention Transformer for Semi-Supervised Medical Image Segmentation
\thanks{* These authors contribute equally to this work.}
\thanks{$\dag$ Corresponding authors.}
}
%


\author{
\IEEEauthorblockN{
Fuchen Zheng\textsuperscript{12*} \qquad Quanjun Li\textsuperscript{3*} \qquad Weixuan Li\textsuperscript{3} \qquad Xuhang Chen\textsuperscript{12} \qquad Yihang Dong\textsuperscript{2} \qquad Guoheng Huang\textsuperscript{3} \\ Chi-Man Pun\textsuperscript{1$\dag$} \qquad Shoujun Zhou\textsuperscript{2$\dag$}
}
\IEEEauthorblockA{
\textsuperscript{1}University of Macau \\
\textsuperscript{2}Shenzhen Institute of Advanced Technology, Chinese Academy of Sciences \\
\textsuperscript{3}Guangdong University of Technology
}
}

\maketitle

\begin{abstract}
Medical image segmentation, a critical application of semantic segmentation in healthcare, has seen significant advancements through specialized computer vision techniques. While deep learning-based medical image segmentation is essential for assisting in medical diagnosis, the lack of diverse training data causes the long-tail problem. Moreover, most previous hybrid CNN-ViT architectures have limited ability to combine various attentions in different layers of the Convolutional Neural Network.
To address these issues, we propose a Lagrange Duality Consistency (LDC) Loss, integrated with Boundary-Aware Contrastive Loss, as the overall training objective for semi-supervised learning to mitigate the long-tail problem. Additionally, we introduce CMAformer, a novel network that synergizes the strengths of ResUNet and Transformer. The cross-attention block in CMAformer effectively integrates spatial attention and channel attention for multi-scale feature fusion.
Overall, our results indicate that CMAformer, combined with the feature fusion framework and the new consistency loss, demonstrates strong complementarity in semi-supervised learning ensembles. We achieve state-of-the-art results on multiple public medical image datasets. Example code are available at:
\url{https://github.com/lzeeorno/Lagrange-Duality-and-CMAformer}.
\end{abstract}



\section{Introduction}
Artificial intelligence-based medical image segmentation has be-
come a crucial tool in clinical therapy, particularly for early cancer
detection and prevention~\cite{1}. However, medical imaging presents
unique challenges compared to natural image processing, primarily
due to the complexity of high-resolution 3D structures and the
substantial costs associated with image acquisition. To be more specific,
the scarcity of expert annotators, often constrained by demanding
schedules, poses additional difficulties in obtaining accurately labeled
datasets. Therefore, the long-tail problem~\cite{longTail2023deep,32long-tail} has become one of the biggest challenges in deep learning-assisted medical image analysis. 

On the other hand, U-Net architecture has been instrumental in advancing seg-
mentation technology, with extensions like ResUnet~\cite{6resunet} combining
the strengths of U-Net~\cite{4unet} and ResNet~\cite{7resnet}. ResUnet incorporates
residual connections to mitigate vanishing gradients~\cite{7resnet,8residual_connection,9res_learning}, enabling
the training of deeper networks and enhancing information flow for
more precise predictions. Additionally, it leverages U-Net’s skip con-
nections to merge feature maps from various resolutions, improving
the model’s ability to capture fine details while maintaining global
context.
Recent developments have seen the integration of transformer
models, known for their robust feature extraction capabilities, with
ResNet to create advanced architectures such as ResT~\cite{resT2021v1, resT2022v2}. However, the potential of combining multi-attentions with hybird CNN-Transformer model remains
largely unexplored. 

The main contributions of this paper are as follows:
\begin{enumerate}
    \item We propose \textbf{Compound Multi-Attention Transformer (CMAformer)}, a novel model that leverages the respective strengths of ResUnet and Transformer, while fusion the spatial attention and channel attention in multi-scale by the proposed \textbf{Cross Attention} layer. 

    \item To address long-tail problem in medical image analysis area, we propose a semi-supervised learning Framework by propose a \textbf{Lagrange Duality Consistency (LDC) Loss} which utilizing Lagrange multipliers reformulate BCE-Dice loss function as a convex optimization consistency loss.

    \item Through a series of experiments on numerous public medical image datasets, CMAformer surpasses the majority of state-of-the-art models in segmentation tasks.
\end{enumerate}

\section{Related Work}
\subsection{CNNs and ViTs}
The U-Net architecture's effectiveness in image segmentation stems from its upsampling process, which restores abstract features to match the original image size, and its downsampling, which enhances the receptive field. However, the original U-Net lacks information exchange among different layers. To address this limitation, Zhou \etal introduced UNet++ \cite{13unet++2018}, which incorporates intermediate nodes within each layer and establishes long connections through feature concatenation, thereby improving information sharing and integration.

The integration of attention mechanisms into U-Net architectures has led to the development of U-shape Transformer. Oktay \etal \cite{14attUnet2018} demonstrated that attention units enable various parts of the network to focus on segmenting multiple objects simultaneously. Building on this concept, Chen \etal \cite{15chen2021transunet} proposed TransUNet, which combines a convolutional network for feature extraction with a transformer for encoding global context. Similar approaches \cite{16transfuse2021,17medicaltransformer,18chang2021transclaw,19chen2023transattunet} have emerged, yet they often underutilize the transformer component, employing only a limited number of layers. Consequently, these models fail to fully incorporate long-term dependencies into convolutional representations. The optimal integration of convolution and self-attention for medical image segmentation remains an open research question.

\subsection{Semi-Supervised Learning}
Traditional supervised learning methods heavily rely on labeled data, which can be resource-intensive and sometimes infeasible to obtain. Semi-supervised learning \cite{22semiSupervised2009} addresses this challenge by leveraging both labeled and unlabeled data to enhance model generalization and performance.

Recent advancements in semi-supervised segmentation have primarily focused on two approaches. The first involves collaborative training \cite{23co-training} and self-training \cite{24self-training} methods, which utilize pseudo labels to incorporate unlabeled data into the training process. The second approach emphasizes the regularization \cite{26regularization_ssl} and optimization \cite{27optimization} of consistency loss functions \cite{28jeong2019consistency,29consistency_ssl2020}. These techniques aim to improve segmentation accuracy by enforcing consistency between predictions on unlabeled samples.

This study builds upon these advancements by optimizing the consistency loss function in conjunction with collaborative training strategies for multi-task segmentation networks. By combining these techniques, we aim to further enhance the accuracy and generalization capabilities of semi-supervised segmentation models in medical imaging applications.

\begin{figure}[h]
\centerline{\includegraphics[width=\columnwidth]{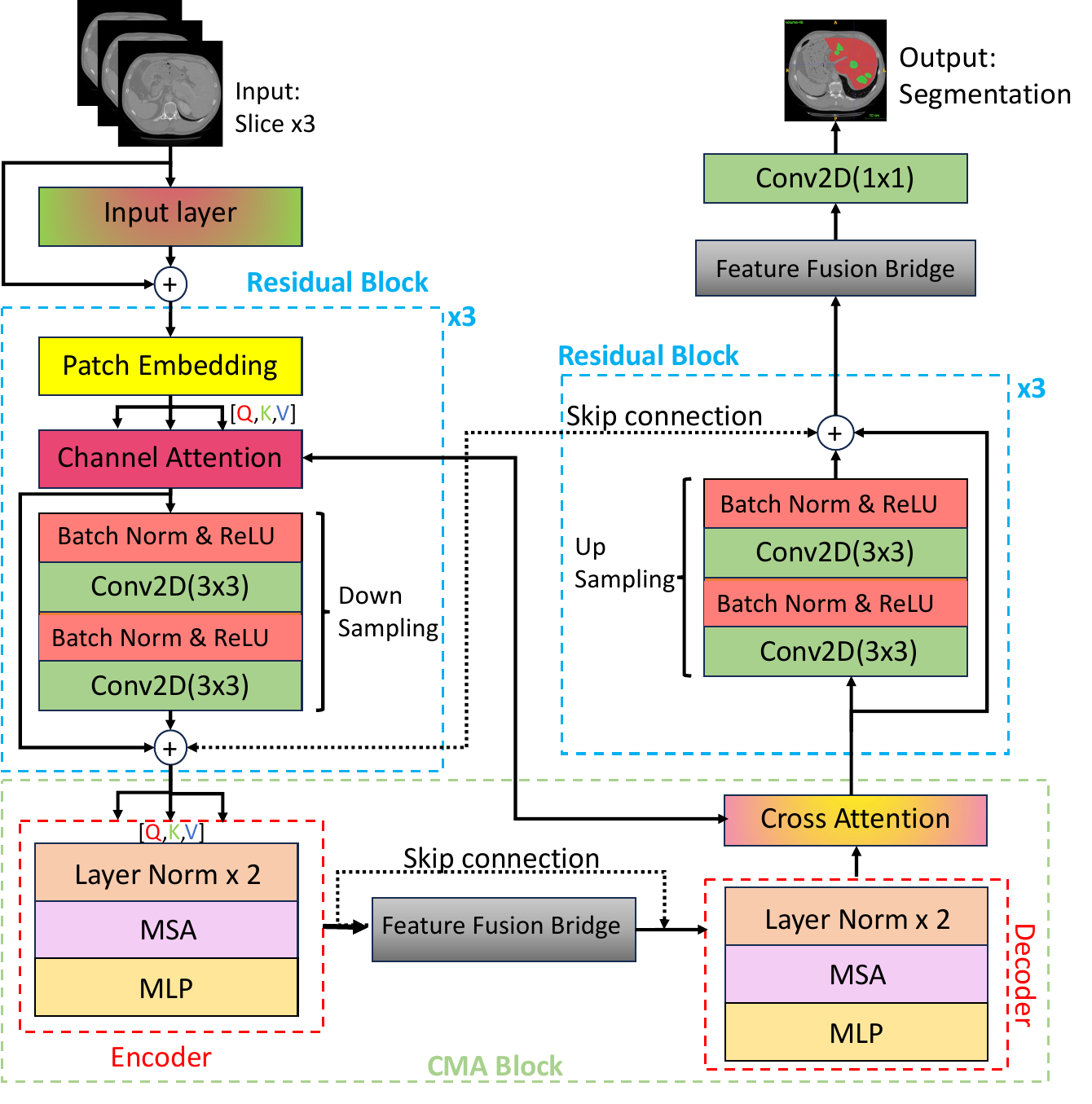}}
\caption{The proposed CMAformer architecture.}
\label{fig2}
\end{figure}

\section{Method}
In this section, we detail the architecture of our CMAformer. Additionally, we introduce a semi-supervised learning framework for robust lesion segmentation that we have proposed.
This framework intrgrated a Lagrange Duality Consistency (LDC) Loss proposed by us and an unsupervised boundary-aware contrastive objective function~\cite{54contrastive2020contrastive,55simcvd2022simcvd} so that we can utilize a large volume of unlabeled data to improve the model's ability to detect both common and rare lesions.



\subsection{Channel Attention and Cross Attention}

CMAformer employs residual blocks as its basic building unit. These blocks, while maintaining the conventional structure of two convolutional layers and a skip connection, incorporate unique enhancements inspired by transformers and ResU-shaped networks.

First, the feature map undergoes patch embedding within the residual block and is divided into $p\times p$ patches.To preserve positional information crucial for spatial understanding, we map the 2D coordinates of each patch onto its flattened representation. However, downsampling operations within the network alter the feature map's shape, rendering the initial position embedding ineffective. To address this, we utilize channel attention, inspired by SEnet \cite{33SEnet}, to remap the positional information from the patches onto the feature map's channels.



This combined approach of patch embedding and channel attention allows CMAformer to effectively fuse spatial and channel-wise information, enhancing its ability to learn discriminative features. The skip connection within the residual block further improves information flow by adding features extracted from the main branch after convolution and batch normalization, promoting gradient propagation and facilitating training. Finally, the main feature map undergoes downsampling through two convolutional layers and ReLU activation, preparing it for subsequent processing stages.

The CMAformer's cross-attention layer after the transformer decoder, the, the input feature map is initially split into two components: the query and the key-value pair. The query typically stems from the current layer's output, whereas the key-value pair may originate from various channels within the same layer or from feature maps across different layers. Initially, we project the query $(Q)$ and the key-value $(K, V)$ feature maps into a unified feature space via linear transformations. Subsequently, we calculate the similarity between the query and the key to determine the attention weights. A weighted summation of the values, guided by these attention weights, yields the integrated feature representation. The cross-attention mechanism allows CMAformer to dynamically amalgamate information from disparate feature maps, significantly boosting its capability to represent features. When integrated with channel attention, this mechanism empowers CMAformer to assimilate comprehensive feature details across multiple scales and contexts, thus elevating its performance in complex visual tasks.

\begin{table*}[h]
    \centering
    \caption{Comparison with State-of-the-Art models on the Synapse multi-organ Dataset. The best results are bolded while the second best are underlined.}
    \resizebox{\textwidth}{!}{%
    \begin{tabular}{c|c|c|c|c|c|c|c|c|c}
        \toprule
        Model  & Average & Aotra & Gallbladder & Kidney(Left) & Kidney(Right) & Liver & Pancreas & Spleen & Stomach \\
        \midrule

        ViT\cite{34vit}+CUP\cite{15chen2021transunet}    
            & 67.86 & 70.19 & 45.10 & 74.70 & 67.40 & 91.32 & 42.00 & 81.75 & 70.44     \\
            
        TransUNet\cite{15chen2021transunet}
            & 84.36 & 90.68 & \underline{71.99} & 86.04 & 83.71 & 95.54 & 73.96 & 88.80 & 84.20      \\
    
        SwinUNet\cite{21swinUnet2022}
            & 79.13 & 85.47 & 66.53 & 83.28 & 79.61 & 94.29 & 56.58 & 90.66 & 76.60    \\
    
        Swin UNETR\cite{49swinUNETR}
            & 83.51 & 90.75 & 66.72 & 86.51 & 85.88 & 95.33 & 70.07 & \textbf{94.59} & 78.20     \\

        TransClaw U-Net\cite{18chang2021transclaw}
            & 78.09 & 85.87 & 61.38 & 84.83 & 79.36 & 94.28 & 57.65 & 87.74 & 73.55    \\

        LeViT-UNet-384s\cite{50levit}
            & 78.53 & 87.33 & 62.23 & 84.61 & 80.25 & 93.11 & 59.07 & 88.86 & 72.76     \\

        CoTr\cite{52cotr}
            & 80.78 & 85.42 & 68.93 & 85.45 & 83.62 & 93.89 & 63.77 & 88.58 & 76.23     \\

        UNETR\cite{48unetr}
            & 79.56 & 89.99 & 60.56 & 85.66 & 84.80 & 94.46 & 59.25 & 87.81 & 73.99     \\

        nnFormer\cite{47nnformer}
            &\underline{86.57} & \underline{92.04} & 70.17 & \textbf{86.57} & \underline{86.25} & \underline{96.84} & \underline{83.35} & 90.51 & \underline{86.83}     \\
        \midrule
        \textbf{CMAformer(Ours)}
            & \textbf{87.39} & \textbf{93.21} & \textbf{72.03} & \underline{86.55} & \textbf{86.62} & \textbf{97.78} & \textbf{83.81} & \underline{92.19} & \textbf{86.92}     \\
        \bottomrule
    \end{tabular}
    }
    \label{tab:syna}
\end{table*}

\subsection{Transformer Block and Spatial Attention Block}

\begin{figure}[h]
\centerline{\includegraphics[width=.8\columnwidth]{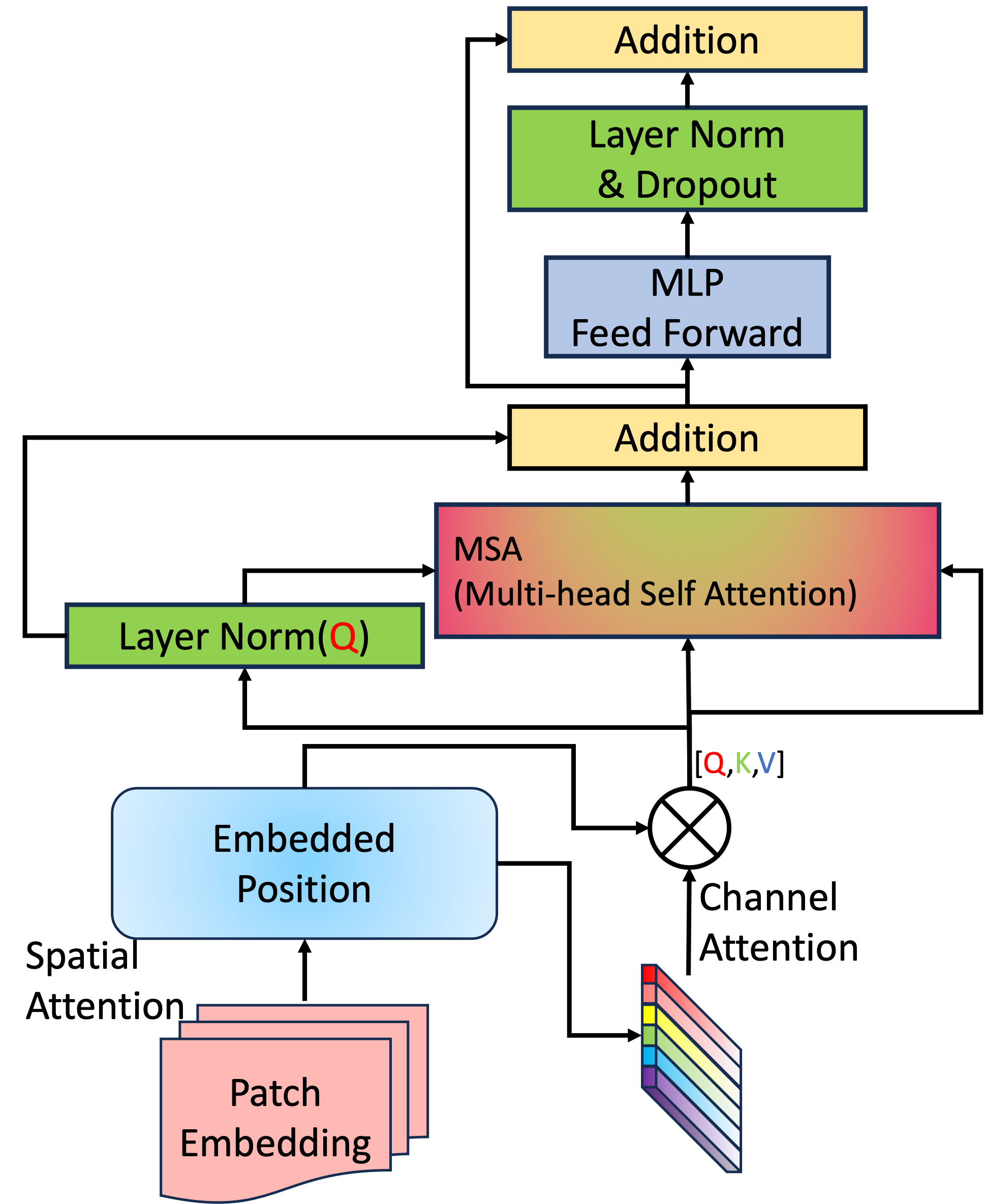}}
\caption{The proposed CMA Block.}
\label{fig3}
\end{figure}

Each transformer block consists of multi-head self-attention (MSA) and a multi-layer perceptron (MLP). To leverage the strengths of transformers for image data, we introduce two key modifications.

\begin{figure}[h]
\centerline{\includegraphics[width=1\columnwidth]{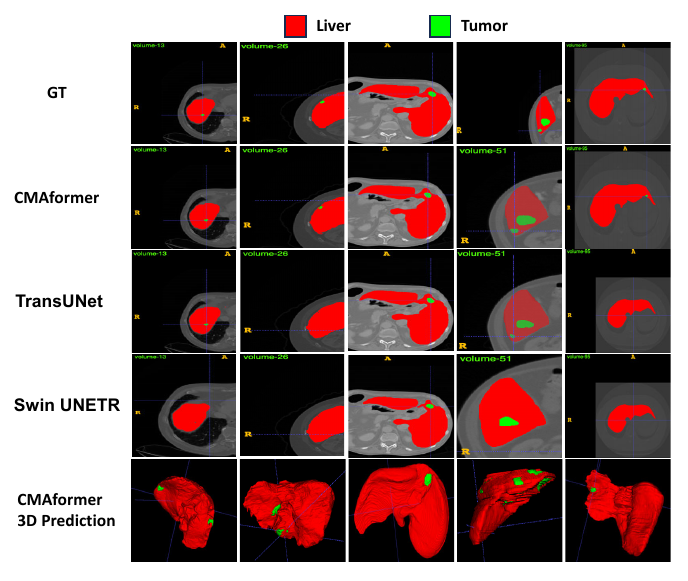}}
\caption{LiTS2017 Segmentation Result}
\label{fig6}
\end{figure}

First, instead of layer normalizing the entire output of the MSA, we perform layer normalization individually on the query matrix $Q$. This modification promotes a more stable gradient signal during training, improving optimization efficiency. Second, the skip connection within the transformer block directly adds the query matrix $Q$ to the MSA output before feeding it to the MLP. This design minimizes information loss and allows the transformer to better model long-range dependencies.
To effectively handle the multi-scale nature of medical images, Transformer Block incorporates a spatial attention block between the encoder and decoder. Inspired by atrous spatial pyramid pooling (ASPP) \cite{41ASPP}, this block fuses feature maps from different scales using dilated convolutions. By integrating these components, Transformer Block effectively model long-range dependencies, while capture local information.

Before inputting the feature map to the transformer block, it is reshaped to match the expected input shape of $[B, H\times W, C]$, where $B$ is the batch size. We utilize the GELU activation function within the MLP due to its smoother nature compared to ReLU. The MLP calculation follows the standard formulation:
\begin{equation}
\operatorname{MSA}(Q, K, V) = \operatorname{softmax}\left(\frac{QK^T}{\sqrt{d_k}} + P\right)V,
\end{equation}
where $P\in\mathbb{R}^{N_T}$ represents the relative position encoding and $d_k$ is the dimensionality of the key vectors. While the computational complexity remains similar to basic ViTs \cite{34vit}, the structural adjustments in Transformer Block significantly enhance its suitability for medical segmentation tasks.

\subsection{Lagrange Duality Consistency (LDC) Loss}
To address the challenges of complex medical image segmentation tasks, we propose a modified BCE-Dice loss function as the consistency loss \cite{39consistencyloss} and reformulate it as a convex optimization problem using Lagrangian duality. The consistency loss is defined as:
\begin{equation}
\mathcal{L}_{con} = -\frac{1}{N} \sum[\alpha y_i \log(p_i) + (\beta_1\frac{y_{i1}p_{i1}}{y_{i1} + p_{i1}} + \beta_2\frac{y_{i2}p_{i2}}{y_{i2} + p_{i2}})],
\end{equation}
where $\alpha$, $\beta_1$, and $\beta_2$ are weighting factors, with $\beta_1$ and $\beta_2$ set to $[0.314 \pm 1e-5, 0.685 \pm 1e-5]$ empirically.

To optimize this loss function, we introduce Lagrange multipliers to handle various constraints: (1) Upper and lower bounds on output values. (2) Limitation on the sum of output values. (3) $L_2$ norm limit on the model parameter vector. (4) Non-negativity constraints.

These constraints are incorporated into a Lagrangian function:
\begin{equation}    
    \begin{split}
    \mathcal{L}(p, w, \mu^+, \mu^-, \lambda, \eta, \zeta) =
    \sum_{i=1}^N (p_i - \mu_i^-) +  (\mu_i^+ - p_i) \\
    +\lambda (\sum_{i=1}^N p_i - S) 
    +\eta ( C - |w|_2^2) 
    +\sum_{i=1}^N \zeta_i (-p_i) ,    
    \end{split}
\end{equation}
This approach, solved using Karush-Kuhn-Tucker (KKT) conditions, allows for more effective optimization of the consistency loss, particularly improving the segmentation of small objects like tumors in complex medical imaging tasks.

\subsection{Overall Training Objective}
Inspired by recent work~\cite{54contrastive2020contrastive,55simcvd2022simcvd}, we cited an unsupervised boundary-aware contrastive objective function. This approach utilizes the InfoNCE loss~\cite{56InfoNCE2018representation} to contrast positive and negative samples by leveraging "boundary-aware" knowledge. The contrastive learning objective ensures consistency between predicted labeled and unlabeled outputs during training as follow:
\begin{equation}
    \mathcal{L}_{contrast}(h_{i,j}^{t}, h_{i,j}^{s}) = -\log \frac{\exp\left(h_{i,j}^{t} \cdot h_{i,j}^{s}/{\tau}\right)}{\sum_{k,l} \exp\left(h_{i,j}^{t} \cdot h_{k,l}^{s}/{\tau}\right)} ,
\end{equation}
where $h_{i,j}^{t}$ and $h_{i,j}^{s}$ represent the projection heads of the teacher and student's signed distance map (SDM)~\cite{58SDM_old2015motion,57SDM2019distance}, respectively. The term $\tau$ denotes a temperature hyperparameter. The indices $k$ and $l$, used in the denominator, are randomly selected from a mini-batch of images, where $i$ and $j$ refer to the CT sample index and the slice index, respectively. The term $h_{k,l}^{s}$ refers to the negative samples.
Therefore, the overall loss function is:
\begin{equation}
    \mathcal{L}_{total} = \lambda \cdot \mathcal{L}_{sup} + \beta \cdot \mathcal{L}_{contrast} + \gamma \cdot \mathcal{L}_{con} ,
\end{equation}
Here, $\mathcal{L}_{sup}$ represents the standard MSE loss~\cite{59MSE1986learning} for labeled data, $\mathcal{L}_{contrast}$ represents the unsupervised boundary-aware contrastive objective function, and $\mathcal{L}_{con}$ denotes the LDC loss proposed by us. The hyperparameters $\lambda$, $\beta$, and $\gamma$ are used to balance each term.

\section{Experiments}
To compare CMAformer fairly with previous models, experiments were conducted on two datasets: the liver tumor segmentation task in LiTS2017 \cite{44lits2017} and the multi-organ segmentation task in Synapse \cite{45synapse}.

\subsection{Implementation Details}
For most experimental settings, the instructions of nnformer \cite{47nnformer} and UNETR \cite{48unetr} were followed. The datasets are split at 80\%, 15\%, and 5\% to form the training/validation/test sets, respectively. We run all experiments based on pytorch framework on Ubuntu 22.04. All training procedures have been performed on a single NVIDIA GeForce RTX 4090 GPU. The initial learning rate is set to 0.001, and we apply a ``poly'' decay strategy. The default optimizer is SGD, where we set the momentum to 0.99. The weight decay is set to $3e-5$.

\subsection{Comparison with State-of-the-Art Models}
\begin{figure}[h]
\centerline{\includegraphics[width=1\columnwidth]{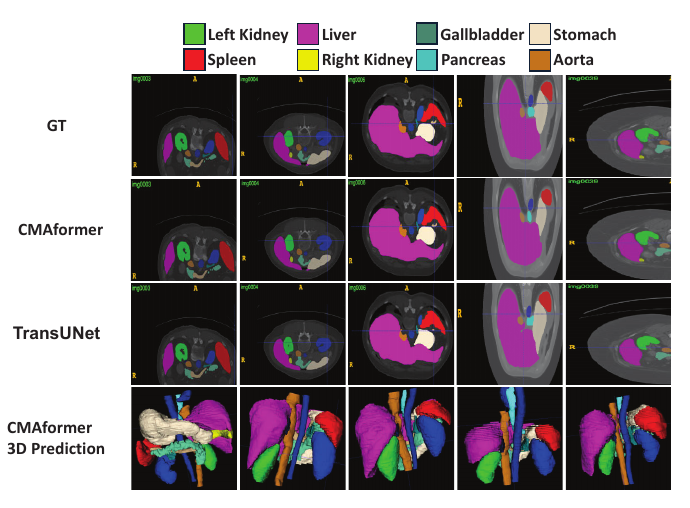}}
\caption{Synapse Segmentation Result}
\label{fig7}
\end{figure}

 Table~\ref{tab:syna} shows the Synapse results. Although CMAformer did not surpass previous models in all metrics, the average of about 10 rounds exceeded all previous advanced models. Table~\ref{tab:lits} presents the results on the LiTS2017 dataset. CMAformer demonstrates robust performance in medical image segmentation, surpassing prior models across several metrics. Within the semi-supervised learning framework, utilizing only 50\% of labeled data, CMAformer achieves tumor segmentation results suppress those of leading-edge methods. Figure.\ref{fig6} and Figure.\ref{fig7} are the segmentation results. 

The two pictures correspond to two datasets respectively. It can be found that our method is better at segmenting small objects such as tumors. We also include 3D reconstruction renderings for these two datasets. 3D reconstruction is mainly implemented based on stacking of predicted 2D images.

Overall, CMAformer achieved competitive and often superior performance compared to previous models across the multiple medical image segmentation datasets under the proposed semi-supervised learning framework. Our LDC loss strategy also helped CMAformer achieve strong results for the semi-supervised tasks.

\begin{table}[h]
    \centering
    \caption{Comparison with State-of-the-Art models on the LITS2017 Dataset. The ``SSL'' suffix indicates that semi-supervised learning utilizes only 50\% of the labeled data. The best results are bolded while the second best are underlined.}
    \begin{tabular}{c|c|c|c}
        \toprule
        Model  & Average & Liver  & Tumor \\
               
        
        \midrule
        ResUnet++ \cite{53resunet++}
            & 82.62 & 85.83 & 79.41     \\
        
        ResT-V2-B \cite{resT2022v2}
            & 90.91 & 94.88 & 86.93     \\
        
        TransUNet \cite{15chen2021transunet}
            & 90.65 & 94.56 & 86.73      \\
    
           
       Swin UNETR\cite{49swinUNETR}
           & 94.41 & \underline{97.10} & 91.71     \\

        nnFormer \cite{47nnformer}
            & 93.10  & 96.01 & 90.16     \\
        
        \midrule
        \textbf{CMAformer-SSL(Ours)}
            & \underline{95.27} & 96.32 & \textbf{94.21}    \\
       \textbf{CMAformer(Ours)}
           & \textbf{95.62} & \textbf{97.89} & \underline{93.34}     \\
        \bottomrule
    \end{tabular}
    \label{tab:lits}
\end{table}

\subsection{Ablation Study}


\begin{table}[h]
\centering
\caption{Ablation Study Result On The LITS2017 Dataset}
\begin{tabular}{ c c c | c c c}
    \hline
     ViT Block & Cross Attention & LDC Loss & Average   & Liver   & Tumor   \\
    \hline
    {\checkmark} & $\times$ & $\times$ &83.29  &85.64  &80.93  \\
    {\checkmark} & {\checkmark} & $\times$ & 92.44 & 95.94 & 88.94 \\
    $\times$ & $\times$ & {\checkmark} & 93.50 & 96.57 & 90.43 \\
    {\checkmark}&{\checkmark}&{\checkmark}&95.62 &97.89 &94.21 \\
    \hline
\end{tabular}
\label{table5}
\end{table}
The outcomes of the ablation experiments are presented in Table~\ref{table5}, highlighting three critical components of our framework: LDC Loss, ViT Block, and Cross Attention layer. The results indicate that the application of a semi-supervised learning framework enhances performance by approximately 12.25\%, with even more substantial improvements observed in tumor segmentation. Furthermore, the integration of the Cross Attention layer with the ViT block significantly boosts performance by about 10.98\%, underscoring the efficacy of the proposed components.




\section{Conclusion}
This paper proposed CMAformer, a hybird model that combines the strengths of ResUNet and Transformer, for medical image segmentation. Vit block in CMAformer utlize our proposed Cross Attention layer achieved modeling long-range dependencies while maintaining ability to capture local feature. Moreover, a semi-supervised learning framework based on Lagrange Duality Consistency Loss was developed to maximize the performance of CMAformer and try to address the long-tail problem.
The experimental results for different segmentation tasks demonstrate the effectiveness and generalizability of CMAformer compared to previous models. The rigorous experimental setup and statistical analysis provide insights into CMAformer's performance gains.
The ability to accurately segment small lesions from medical images can benefit downstream tasks for computer-aided diagnosis and treatment planning. Future work will explore applying CMAformer to more medical image segmentation tasks and real-world clinical applications.

\newpage
\bibliographystyle{IEEEtran}
\bibliography{ref}

\end{document}